# Swarm Intelligence in Collision-free Formation Control for Multi-UAV Systems with 3D Obstacle Avoidance Maneuvers

Reza Ahmadvand[1], Sarah Safura Sharif[1], Yaser Mike Banad[1*]

[1]School of Electrical and Computer Engineering, University of Oklahoma, Oklahoma, 73019, United States

([*]Corresponding author's email: bana@ou.edu)

**ABSTRACT**
Recent advances in multi-agent systems manipulation have demonstrated a rising demand for the implementation of multi-UAV systems in urban areas which are always subjected to the presence of static and dynamic obstacles. The focus of the presented research is on the introduction of a nature-inspired collision-free control for a multi-UAV system considering obstacle avoidance maneuvers. Inspired by the collective behavior of tilapia fish and pigeon, the presented framework in this study uses a centralized controller for the optimal formation control/recovery, which is defined by probabilistic Lloyd's algorithm, while it uses a distributed controller for the intervehicle collision and obstacle avoidance. Further, the presented framework has been extended to the 3D space with 3D maneuvers. Finally, the presented framework has been applied to a multi-UAV system in 2D and 3D scenarios, and obtained results demonstrated the validity of the presented method in the presence of buildings and different types of obstacles.

## INTRODUCTION

Swarm intelligence, an emergent property observed in nature, has long served as a source of inspiration for engineering applications, particularly in the development of autonomous systems [1]. This phenomenon demonstrates how decentralized systems, composed of numerous simple agents, can achieve complex collective behaviors without a central controller. In recent years, multi-agent systems (MAS) have gained significant attention in both civil and military applications, including intelligent transportation systems, surveillance, and search-and-rescue operations [2]. These applications demand sophisticated control strategies that can ensure both mission effectiveness and operational safety in complex urban environments, where complex obstacles, such as buildings and both static and dynamic obstructions, are prevalent.

While traditional research has focused primarily on improving system accuracy and autonomy, safety has become increasingly critical. In urban environments, ensuring that autonomous agents, such as unmanned aerial vehicles (UAVs), can operate safely and efficiently without colliding with obstacles or each other is critical. Although extensive literature exists on the traditional centralized control methods for multi-UAV systems, their reliability and mission success heavily depend on the central computer—a significant constraint that can lead to complete mission failure if the central computer malfunctions [2, 3].

The primary focus of our study is to introduce a nature-inspired control method for the collision-free formation control of multi-UAV systems operating in environments containing buildings, static obstacles, and dynamic obstacles, utilizing 3D obstacle avoidance maneuvers. Previous research on obstacle avoidance has identified several approaches: path planning and geometric guidance methods, which require a known environmental map with modeled obstacles for the optimal collision-free path between two desired points [4, 5]; potential field function approaches based on defined potential field functions [6]; and model predictive-based control approaches that make decisions about control inputs based on predicted system dynamics [7, 8]. Recent advances in the multi-UAV control systems show that stand-alone path planning methods and their combinations with methods like particle swarm optimization (PSO) remain active areas of research [9, 10]. Furthermore, strategies based on the artificial potential field (APF) concepts have been introduced for implementing multi-UAV systems in environments with known obstacle maps [11, 12, 13, 14].

In [15], researchers statistically investigated pigeons' self-organized obstacle avoidance behavior and developed a mathematical model for their obstacle avoidance maneuvers. This pigeon-inspired avoidance model was implemented for a group of four quadrotors [16]. Although numerous methods exist for static obstacle avoidance, dynamic obstacle avoidance remains an open problem [3]. A novel pigeon-inspired method was proposed in [3] for dynamic obstacle avoidance, addressing formation control and recovery through consensus theory. However, this approach increased controller design complexity and limited obstacle detection and avoidance to planar maneuvers, sometimes resulting in UAVs becoming constrained between neighboring UAVs and external obstacles. The problem of 3D obstacle avoidance maneuvers has been primarily investigated for single UAV problems. In [17], researchers introduced a path-planning approach for single UAV problems utilizing multi-objective spherical vector-based PSO. Moreover, [18] introduced a novel 3D route planning method for single UAVs by leveraging hybridized slime mould with a different updating algorithm and employing the Pareto optimality.

Drawing inspiration from research on 3D maneuvers and the pigeon-inspired method introduced in [3], our study proposes a novel approach for simultaneous consideration of formation control, inter-vehicle collision avoidance, and obstacle avoidance in the presence of static and dynamic obstacles for multi-UAV systems. In this approach, optimal UAV positioning within formations is inspired by territorial behavior observed in tilapia fish. This behavior ensures that the UAVs maintain proper spacing and orientation relative to one another, akin to how fish in a school or birds in a flock exhibit coordinated movement patterns.

Additionally, the obstacle avoidance behavior of pigeons serves as the foundation for maneuvering around static and dynamic obstacles. Pigeons, known for their exceptional navigation and obstacle avoidance abilities, offer a biological model for UAVs to follow, enabling them to detect and circumvent obstacles in real-time. The distributed nature of this control strategy allows each UAV to operate autonomously, with local decision-making capabilities that reduce the reliance on a central controller and enhance overall system robustness.

The primary contributions of this study are: 1) Development of a semi-centralized formation controller inspired by tilapia fish and utilization of the probabilistic Lloyd algorithm [19] in 3D space for optimal configuration and formation change/recovery. 2) Improvement of pigeon-inspired obstacle detection and avoidance through implementation with 3D maneuvers in the presence of dynamic obstacles. 3) Implementation of 3D obstacle avoidance maneuvers in a multi-UAV system consisting of 12 UAVs, significantly expanding upon previous studies that examined only 4 and 6 agents, thus demonstrating our method's scalability for deploying any number of UAVs in the 3D space.

This paper is organized as follows. Section 2 begins with multi-UAV dynamic modeling followed by preliminaries of graph theory, then presents the controller design and theoretical developments. Section 3 presents simulation results for different case studies. Finally, Section 4 concludes the paper and suggests potential applications and future work directions.

## THEORY

This section presents the dynamical model of the multi-UAV system, followed by the essential preliminaries related to graph theory.

### Multi-UAV Dynamical Model

Figure 1 schematically depicts a multi-UAV system consisting of $N$ UAVs and demonstrates the two types of coordinate frames utilized in this study. The first is the inertial coordinate frame described by $O_I X_I Y_I$ used for modeling the translational motion of UAVs relative to the ground surface. The second consists of body coordinate frames described by $O_b x_b y_b$ located at each UAV's center of gravity for consideration of individual flight directions and obstacle locations.

In Figure 1, $\boldsymbol{p}_i$ and $\boldsymbol{v}_i$ represent the position and velocity vectors of the $i^{th}$ UAV in the inertial coordinate frame. $\boldsymbol{p}_{ij} = \boldsymbol{p}_j - \boldsymbol{p}_i$ and $\boldsymbol{v}_{ij} = \boldsymbol{v}_j - \boldsymbol{v}_i$ refer to the relative position and velocity vectors of the $i^{th}$ UAV with respect to the $j^{th}$ UAV. The parameters $r_s$ and $r_d$ denote the safety range and detection range respectively, while $\theta_{FOV}$ refers to the UAV's field of view relative to the flight direction. Following common practice in formation controller design and obstacle avoidance maneuvers, we model the UAVs as particles [3, 20].

The system dynamics are described by:
$$\dot{\boldsymbol{p}}_i = \boldsymbol{v}_i \qquad (1)$$
$$\dot{\boldsymbol{v}}_i = \boldsymbol{u}_i \qquad (2)$$

where $\boldsymbol{u}_i$ refers to the auxiliary control input for the $i^{th}$ UAV.

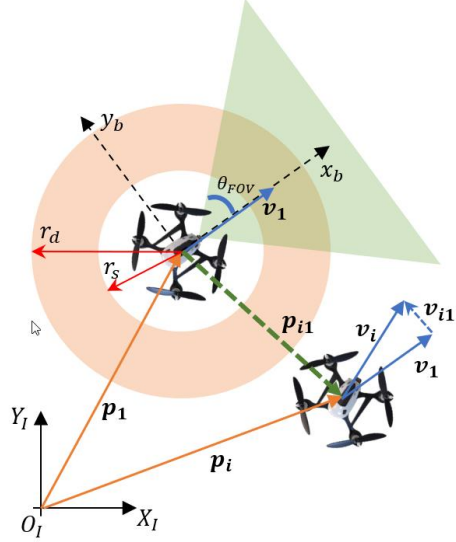

**Figure 1.** Multi-UAV system dynamic model schematic

**Graph Theory**

To provide a mathematical foundation for our controller design, we employ graph theory. In this framework, a multi-UAV system consisting of *n* homogeneous UAVs is described using $G(U, E)$, which represents a graph with a set of nodes defined as $U = \{i\}, i = 1, 2, ..., n$ and a set of edges $E = \{(i, j) | i, j \in U \,;\, i \neq j\}$ that models the communication between the agents.

The set $N_i = \{j \in U \setminus \{i\} | (i, j) \in E\}$ describes the neighborhood for the $i^{th}$ node in the considered graph. Thus, if $\boldsymbol{p}_i$, $i = 1, 2, ..., n$ refers to the position vector of the $i^{th}$ UAV in the inertial coordinate frame and $r_d$ defines its detection range, the UAVs' neighborhoods can be defined using the following set of nodes in space:

$$N_i = \{j \in U \setminus \{i\} \,\big|\, \|\boldsymbol{p}_i - \boldsymbol{p}_j\| < r_d\} \qquad (2)$$

**Formation Controller Design**

This section presents the formation controller design for the optimal configuration of UAVs in a barrier area $Q$. To achieve an optimal formation in a predefined area, the concepts from locational optimization and Voronoi partitions have been leveraged [21, 22, 23].

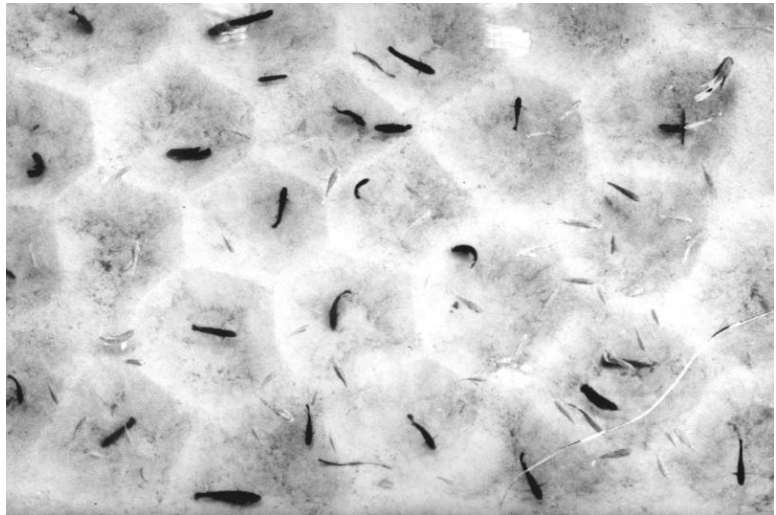

**Figure 2.** Demonstration of Tilapia fish territorial behavior [24]

Drawing inspiration from the tilapia fish territorial behavior, as demonstrated in Figure 2, which can be modeled by a centroidal Voronoi configuration, and building on investigations of the Voronoi configurations in [20], we note that the sensing performance of any UAV at any point $q$ in its sensing range within a desired area $Q$ depends heavily on the distance $||q - p_i||$. To quantify the sensing performance of the system, we consider the following multicenter cost function, which provides a measure of the expected sensing performance by all UAVs at any point $q$ in space:

$$J(p_1, \ldots, p_N) = \int max\, f(||q - p_i||)\phi(q)dq \quad (3)$$

To achieve optimal Voronoi partitioning, we define:

$$V_i = \{q \in Q : ||q - p_i|| \leq ||q - p_j||, j \neq i\} \quad (4)$$

This definition of the Voronoi partitions, combined with the cost function defined in Eq. (4) leads to the following differential equation showing the differentiation of cost function with respect to UAVs location in inertial frame:

$$\frac{\partial J_v}{\partial p_i} = M_{v_i}(c_{v_i} - p_i) \quad (5)$$

This expression demonstrates that the optimal solution for the considered cost function (where $\frac{\partial J_v}{\partial p_i} = 0$) occurs at locations satisfying $p_i = c_{v_i}$ resulting in a centroidal Voronoi tessellation (CVT). Here, $M_{v_i}$ represents the Voronoi center of mass, and $c_{v_i}$ is Voronoi centroid. Further, in our research, we utilize the probabilistic Lloyd's algorithm for CVT implementation [19]. Table 1 presents the pseudo-code of the implemented Lloyd's algorithm, providing a systematic approach to achieving optimal UAV distribution. The algorithm iteratively refines UAV positions through sampling and updating steps until convergence criteria are met.

Finally, to achieve optimal configuration for UAVs in our considered barrier area $Q$, we employ the following control law:

$$u_{f_i} = -K_{p_i}(p_i - c_{v_i}) - K_{v_i} v_{i_{rel}} \quad (6)$$

In this expression, the gain $K_{p_i}$ and $K_{v_i}$ are positive definite and can be tuned through trial and error or systematic design methods. $v_{i_{rel}}$ represents the relative velocity between the UAVs and their desired point on the considered barrier, which moves at the speed of 1 m/s in the $x$ direction. This control law uses the Lloyd's algorithm output for CVT as the desired point for the UAVs to reach. Through this approach, we can optimally deploy any number of UAVs in a desired barrier area for either static or dynamic barriers, provided the barrier is large enough to enclose the UAVs with predefined safety areas.

**Collision Avoidance Controller Design**

The preceding section addresses the inter-vehicle collision avoidance, a crucial factor for multi-UAV systems. Inspired by Hook's law and mass-spiring-damper system, considering relative position and velocities between the vehicles, we implement the control law introduced in [3]:

$$u_{c_{ij}} = -k_{c1} p_c + k_{c2} v_{ij} \quad (7)$$

where:

$$p_c = \frac{1}{(||p_{ij}|| - r_s)^2} \frac{p_{ij}}{||p_{ij}||} \quad (8)$$

In the above equations, $k_{c1}$ and $k_{c2}$ are diagonal positive definite matrices which can be tuned by trial and error.

Table 1. Pseudo-Code of the implemented Lloyd's CVT algorithm

| Probabilistic Generalized Lloyd's Algorithm |
|---|
| **Input**:<br>- Domain $Q$<br>- Density function ρ(x) defined on $Q$<br>- Positive integer $N$ (number of points)<br>- Positive integer q (number of sampling points per iteration)<br>- Constants α$_1$, α$_2$, β$_1$, β$_2$ such that α$_1$ + α$_2$ = 1, β$_1$ + β$_2$ = 1, α$_2$ > 0, β$_2$ > 0<br><br>**Steps:**<br>*1. Initialization:*<br>- Choose an initial set of n points {x$_i$}$^n_{i=1}$ in Ω.<br>- Set iteration counters {j$_i$}$^n_{i=1}$ = 1.<br>*2. Sampling:*<br>- Randomly sample q points {y$_r$}$^r_{r=1}$ in Ω using a Monte Carlo method with ρ(x) as the probability density function.<br>*3. Point Update:*<br>  For each i = 1, 2, ..., n:<br>  - Gather all sampled points y$_r$ closest to x$_i$ (forming the set W$_i$, i.e., the Voronoi region of x$_i$).<br>  - If W$_i$ is empty, do nothing.<br>  - Otherwise:<br>    - Compute the average u$_i$ of the points in W$_i$.<br>    - Update x$_i$:<br>      x$_i$ ← ((α$_1$j$_i$ + β$_1$) / (j$_i$ + 1)) x$_i$ + ((α$_2$j$_i$ + β$_2$) / (j$_i$ + 1)) u$_i$<br>    - Increment j$_i$:<br>      j$_i$ ← j$_i$ + 1<br>*4. Repeat or Terminate:*<br>- Form the new set of points {x$_i$}$^n_{i=1}$.<br>- If $Q$ is a hypersurface, project x$_i$ onto $Q$<br>- Check stopping criteria (e.g., convergence or tolerance).<br>- If criteria are not met, go back to Step 2.<br><br>**Output:**<br>- Final set of points {x$_i$}$^n_{i=1}$. |

## Obstacle Avoidance Controller Design

Inspired by the pigeon collective behavior for their self-organized obstacle avoidance maneuvers, we first present the previously introduced obstacle avoidance controller design for planar 2D maneuvers with moving obstacles constrained to only $y$-direction movement [3]. We then extend the method to 3D space through equation reformulation, addressing the main focus of our research. The approach consists of two key components: obstacle detection and maneuver execution through velocity adjustment.

### *Planar Obstacle Detection (2D)*

Figure 3 schematically depicts the obstacle detection strategy for UAVs with considered a sector-like field of view (FOV), inspired by pigeons' natural visual capabilities [3]. The detection mechanism is governed by two key conditions:

$$||\boldsymbol{p}_i - \boldsymbol{o}_k|| \leq r_d + o_{rk} \qquad (9)$$

$$\left|\operatorname{atan}\left(\frac{o_{yk} - p_{yi}}{o_{xk} - p_{xi}}\right) - d_i^{fly}\right| < \theta_{FOV} \; or \; \left|\operatorname{atan}\left(\frac{o_{zk} - p_{zi}}{o_{xk} - p_{xi}}\right)\right| \leq \theta_{FOV} \qquad (10)$$

where $\boldsymbol{o}_k = [o_{kx}, o_{ky}, o_{kz}]$ is the location of $k^{th}$ obstacle in the space. According to these detection rules, an obstacle is detected when two conditions are met: first, the distance between the $i^{th}$ UAV and $k^{th}$ obstacle falls within the detection range, and second, the angle between the flight direction and obstacle is less than considered FOV.

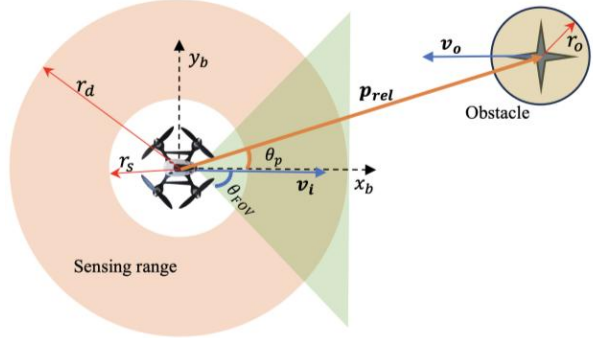

**Figure 3.** Schematic representation of obstacle detection strategy, illustrating the sensing range and FOV.

*Planar Velocity Adjustment (2D)*

Drawing inspiration from pigeons' navigation behavior, where flight direction is determined by considering both obstacle distance and field of view [3], we define a potential function for the avoidance maneuvers as :

$$U_p(\boldsymbol{p}_i, \boldsymbol{o}_k) = \begin{cases} \frac{1}{2}\left(||k_v(\boldsymbol{p}_i - \boldsymbol{o}_k)|| - r_a\right)^2, & detected \\ 0, & otherwise \end{cases} \quad (11)$$

In the above function, $k_v$ is a positive definite diagonal matrix, where it obeys $k_v = diag\{k_x^o, k_y^o\}$, and $r_a = r_d + r_{ok}$. Differentiating the potential function with respect to $p_i$ leads to the following expression:

$$\nabla U_p = \begin{cases} \left(||k_v(\boldsymbol{p}_i - \boldsymbol{o}_k)|| - r_a\right)\nabla(||\boldsymbol{p}_i - \boldsymbol{o}_k||), & detected \\ 0, & otherwise \end{cases} \quad (12)$$

where $\nabla$ refers to the gradient operator, and $\nabla(||\boldsymbol{p}_i - \boldsymbol{o}_k||) = (\boldsymbol{p}_i - \boldsymbol{o}_k)/||\boldsymbol{p}_i - \boldsymbol{o}_k||$.

The above potential function definition could potentially result in UAVs becoming stationary (velocity = 0 m/s) when the forward force equals the evasive obstacle force—effectively creating a local minima. To prevent this condition, we introduce a rotational potential function:

$$U_r(\boldsymbol{p}_i, \boldsymbol{o}_k) = \begin{cases} \frac{k_r}{2}\left(||\boldsymbol{T}_r(\boldsymbol{p}_i - \boldsymbol{o}_k)||\right)^2, & detected \\ 0, & otherwise \end{cases} \quad (13)$$

Differentiation with respect to $\boldsymbol{p}_i$ yields to the following:

$$\nabla U_r = \begin{cases} k_r \boldsymbol{T}_r(||(\boldsymbol{p}_i - \boldsymbol{o}_k)||)\nabla(||\boldsymbol{p}_i - \boldsymbol{o}_k||), & detected \\ 0, & otherwise \end{cases} \quad (14)$$

where, $k_r$ refers to a positive coefficient and $\boldsymbol{T}_r$ is rotation matrix defined as below:

$$\boldsymbol{T}_r = \begin{bmatrix} cos\alpha & -sin\alpha \\ sin\alpha & cos\alpha \end{bmatrix} \quad (15)$$

where:

$$\alpha = \begin{cases} \frac{\pi}{2}\frac{r_d - ||\boldsymbol{p}_{rel_{ik}}||}{r_d - r_s}, & r_s < ||\boldsymbol{p}_{rel_{ik}}|| < r_d \\ 0, & otherwise \end{cases} \quad (16)$$

Here, $\boldsymbol{p}_{rel_{ik}}$ refers to the relative position vector between the $i^{th}$ UAV and $k^{th}$ obstacle in the space. Finally, the obstacle avoidance control law will be achieved as it has been expressed below:

$$\boldsymbol{u}_{oi} = -k_{o1}\sum_{k=1}^{K}\nabla U_p - \sum_{k=1}^{K}\nabla U_r - k_{o2}\boldsymbol{v}_i \qquad (17)$$

where, $k_{o2}$ is a diagonal positive definite matrix which can be tuned using trial and error.

Using the presented control law in Eq. (17), UAVs can perform planar obstacle avoidance in their motion surface. Thus, in the presence of the neighbor UAVs, sometimes it can be challenging for the UAVs to perform their maneuver simultaneously with collision avoidance and they might get stuck for a while between their neighbor UAV and the detected obstacle. To address this limitation and provide additional maneuverability (i.e., additional degrees of freedom) to the UAVs, we extend the method to 3D space using rotational matrices and 3D rotation concepts [25].

### *Nonplanar Obstacle Detection (3D)*
As shown in Figure 4, we modify the previous assumptions to consider spherical sensing and safety ranges around the UAVs, with a conical FOV region around the flight direction.

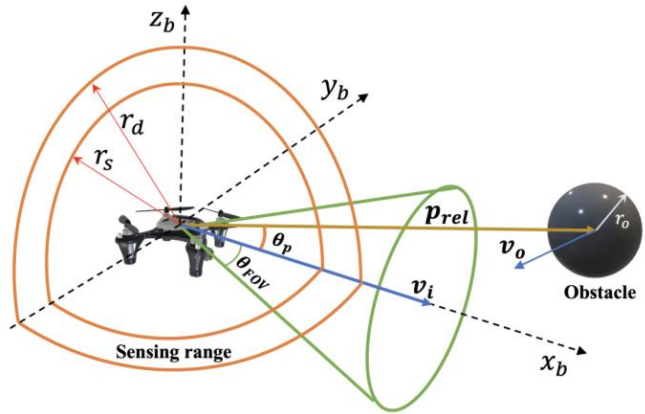

**Figure 4.** Schematic representation of obstacle detection strategy, illustrating the sensing range and FOV in 3D space

Considering the assumptions mentioned, the implementation of vectors inner product properties in the space led to the following set of conditions for the obstacle detection:

$$||\boldsymbol{p}_i - \boldsymbol{o}_k|| \leq r_d + o_{rk} \qquad (18)$$

$$\left|\left(\cos^{-1}\left(\frac{\boldsymbol{p}_{rel}\cdot\boldsymbol{v}_i}{||\boldsymbol{p}_{rel}||\,||\boldsymbol{v}_i||}\right)\right)\right| \leq \theta_{FOV} \qquad (19)$$

An obstacle is detected when both conditions are met simultaneously, indicating the obstacle is within both the sensing range and the UAV's FOV.

### *Nonplanar Velocity Adjustment (3D)*
In our previously presented framework for planar maneuvers, the rotation matrix used in the potential function corresponded to planar rotation in the $O_b x_b y_b$ plane of the body coordinate frame (Figure 3). To extend this pigeon-inspired maneuver into the 3D space, the rotational potential function presented in Eq. (13) needs to be reformulated as below:

$$U_r(\boldsymbol{p}_i, \boldsymbol{o}_k) = \begin{cases} \frac{k_r}{2}\left(||\boldsymbol{T}_{rx}\boldsymbol{T}_{ry}\boldsymbol{T}_{rz}(\boldsymbol{p}_i - \boldsymbol{o}_k)||\right)^2, & detected \\ 0, & otherwise \end{cases} \qquad (20)$$

where:

$$T_{rx} = \begin{bmatrix} 1 & 0 & 0 \\ 0 & 1 & 0 \\ 0 & 0 & 1 \end{bmatrix} \quad (21)$$

$$T_{ry} = \begin{bmatrix} \cos\alpha & 0 & \sin\alpha \\ 0 & 1 & 0 \\ -\sin\alpha & 0 & \cos\alpha \end{bmatrix} \quad (22)$$

$$T_{rz} = \begin{bmatrix} \cos\alpha & -\sin\alpha & 0 \\ \sin\alpha & \cos\alpha & 0 \\ 0 & 0 & 1 \end{bmatrix} \quad (23)$$

In the above expression, $T_{rx}$, $T_{ry}$ and $T_{rz}$ are rotation matrix for the rotations around $x_b$, $y_b$ and $z_b$ respectively. Differentiation with respect to $p_i$ yields to the following:

$$\nabla U_r = \begin{cases} k_r T_{rx} T_{ry} T_{rz} (||(p_i - o_k)||) \nabla(||p_i - o_k||), & detected \\ 0, & otherwise \end{cases} \quad (24)$$

Finally, utilizing the above-introduced 3D rotational potential function and its derivative, the obstacle avoidance rule presented in Eq. (17) can be implemented. It is notable that, to avoid the rotation around the x-axis of the body coordinate frame, and keep the attitude stable during the maneuvers, the rotation matrix for $T_{rx}$ has been considered as an identity matrix.

### *Overall Controller for Multi-UAV System*

Applying the superposition principle to our previously introduced controller designs, the overall control law for each UAV follows:

$$u_i = u_{fi} + u_{ci} + u_{oi} \quad (25)$$

In this expression, the first term ($u_{fi}$) manages formation control, forcing UAVs to follow their desired points obtained from the Lloyd's method with a non-static barrier area. The second ($u_{ci}$) handles inter-vehicle collision avoidance while the third term ($u_{oi}$) manages obstacle avoidance.

## NUMERICAL SIMULATION

To evaluate the validity of the proposed method in this study, three case studies have been investigated. Firstly, to assess the formation change and recovery of the UAV swarm, the proposed control approach has been applied to a scenario of passing between two buildings. Next, for further assessment, the proposed method has been applied to the same scenario with the added static and dynamic obstacles. Finally, the proposed method has been applied to a full 3D problem in the presence of buildings and various types of obstacles.

### Case Study 1: Collision-free Formation Control

This initial study examines the deployment of eight UAVs in a predefined barrier with a 5-meter height from the ground focusing only on formation control and inter-vehicle collision avoidance (without considering the obstacle avoidance term in the control law):

$$u_i = u_{fi} + u_{ci} \quad (26)$$

We conducted simulations using the parameters in Table 2 over 145 seconds with a time step of 0.1 seconds. Notably, the initial position vectors of UAVs have been sampled from the set $\{(x_{0_i}, y_{0_i}, 0) | 0 < x_i < 1, 0 < y_i < 1: i = 1,2,...,8\}$ using a zero mean Gaussian distribution with unit covariance, assuming stationary at $t = 0$s.

Figure 5 shows the simulation results obtained for the collision-free formation control of a UAV swarm using optimal desired locations obtained from Lloyd's algorithm. The scenario involves a moving barrier area where UAVs perform formation change and recovery maneuvers while encountering a narrow space between two buildings—a common situation in urban environments.

Figure 5(A-a) demonstrates the successful initial deployment of UAVs in the barrier area with optimal configuration. The UAVs maintain their formation while moving in the x direction until the first UAV detects the buildings within its sensing range. Upon building detection, the UAVs execute a formation change without inter-vehicle collisions, adopting a new optimal configuration in a smaller barrier area to navigate through the narrow space between buildings. Figure 5(A-b) illustrates this formation change maneuver as the UAVs pass through the buildings with their new formation. Figure 5(A-c) shows the successful recovery of the initial formation.

Figure 5(B) shows the time history of the obtained distance between UAV1 with all the other UAVs throughout the simulation. The results confirm effective collision avoidance, with no UAVs approaching closer than 2 meters to UAV1. For clarity of visualization, we present only the plan view of the plots. (The complete simulation video is available in the supplementary materials.)

These results validate our proposed method's effectiveness in integrating semi-centralized formation control with distributed collision avoidance, achieving safe and reliable formation change and recovery without requiring complex consensus theory-based methods.

**Table 2.** Numerical values utilized in the simulations

| Parameter | Value |
|---|---|
| $r_d (m)$ | 2 |
| $r_s (m)$ | 1 |
| $K_p$ | $diag([3,3,3])$ |
| $K_v$ | $diag([5,5,5])$ |
| $k_v$ | $diag([0.1,0.5,0.1])$ |
| $k\_r$ | 0.5 |
| $k\_o1$ | 5 |
| $k\_o2$ | 1 |
| $v\_obs4 \ (m/s)$ | $[0.2,0.05,0]$ |
| $r_{ok} \ (m/s)$ | 1 |
| $\theta_{FOV} \ (deg)$ | 60 |

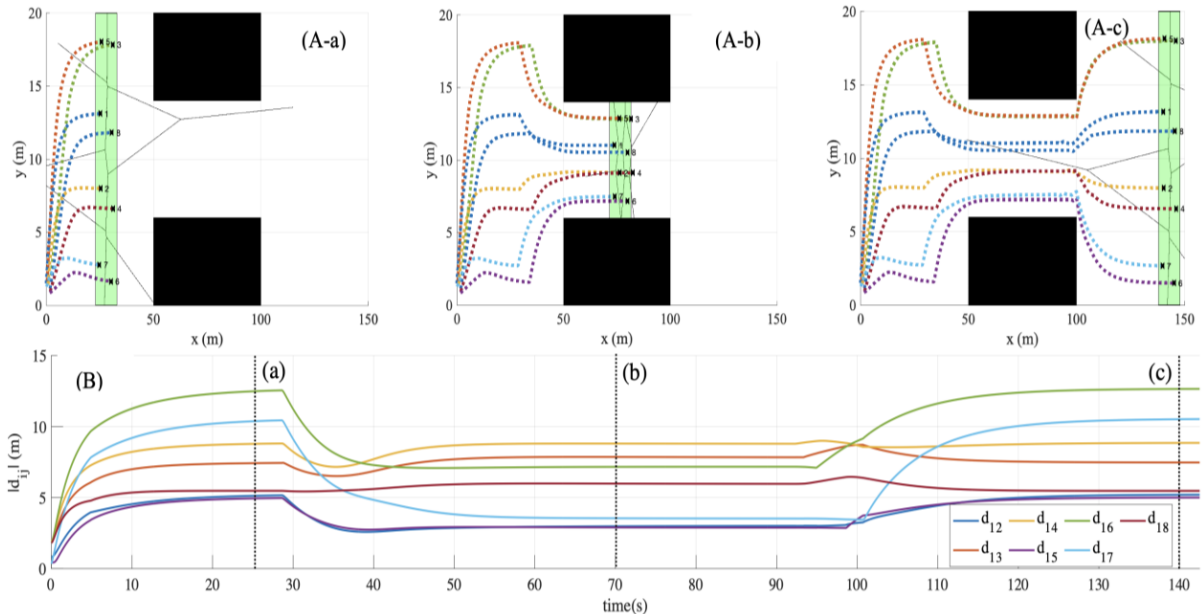

**Figure 5.** simulation results for the collision-free formation control scenario without obstacles

## Case Study 2: Collision-free Formation Control with Obstacle Avoidance

To evaluate the ability of the presented controller in a more complex situation, we applied it to the same formation control scenario but added static and dynamic obstacles. We maintained the same simulation parameters and duration as shown in Table 2. Static obstacles were positioned as shown in the figures, while the dynamic obstacle (obs4) moved with a constant velocity vector of $v_{obs4} = [0.1, 0.025, 0]^T$ m/s, beginning its motion after t = 42s.

Figure 6 illustrates the results of this enhanced scenario. Figure 6(A-a) shows a successful initial UAV deployment, matching the previous case. Figure 6(A-b) depicts the formation change before entering the narrow area between buildings, now complicated by obstacles. At this point, UAVs must simultaneously maintain safe distances from neighboring UAVs, avoid building walls, and execute obstacle avoidance maneuvers.

Figure 6(A-c) demonstrates successful navigation past the static obstacles (obs1, obs2, and obs3) and approach to the dynamic obstacle (obs4). Figure 6(A-d) shows successful passage past obs4 and immediate formation recovery. Figures 6(A-e) and 6(A-f) illustrate the initial and final stages of formation recovery, respectively.

Figure 6(B) presents the time history of distances between UAV1 and other UAVs, with vertical dashed lines corresponding to the situations shown in Figure 6(A). The results confirm maintained safe distances throughout the complex maneuvers. (Complete simulation video available in supplementary materials.)

Obtained results in this section demonstrated the acceptable performance of the proposed method by the integration of previously presented semi-centralized formation controller with distributed collision and obstacle avoidance controllers in a collision-free formation control scenario considering formation change, formation recovery, and obstacle avoidance maneuvers in a safe and reliable way without the need for complex consensus theory-based methods in the presence of static and dynamic obstacles.

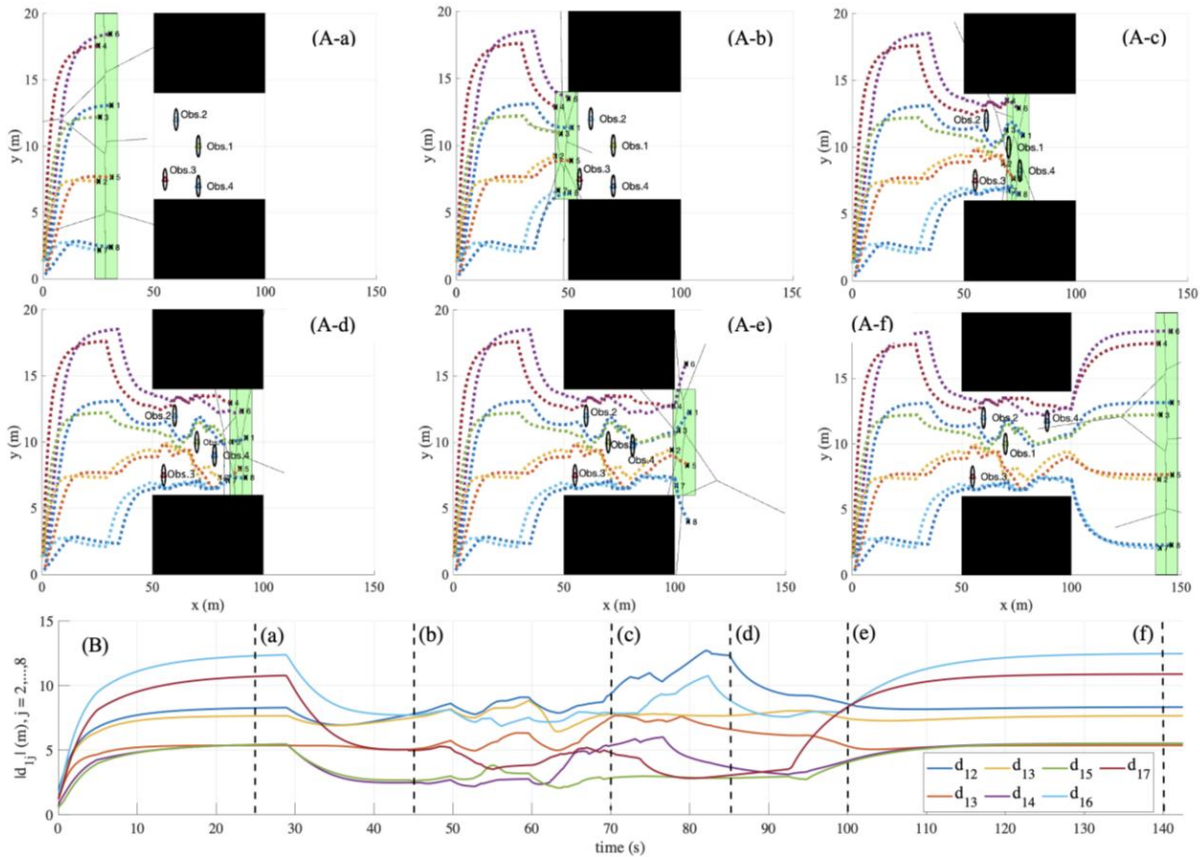

**Figure 6.** Simulation results for collision-free control scenario with planar obstacle avoidance maneuver

## Case Study 3: Collision-free Formation Control with 3D Obstacle Avoidance

This final study evaluates our proposed 3D maneuvers with a larger swarm of 12 UAVs operating in an environment containing buildings, static obstacles, and dynamic obstacles. We conducted simulations over 140 seconds with a 0.1-second time step.

Figure 7 reveals the obtained trajectories in 3D space along with the relative distance history between UAV1 and other UAVs. Figure 7(A-a) demonstrates the successful optimal deployment of UAVs in a 3D barrier space determined by the Lloyd's algorithm, with ground projections of UAV positions clearly shown. Figure 7(A-b) illustrates how the swarm executes formation changes to achieve optimal positioning between buildings while simultaneously performing 3D maneuvers for obstacle avoidance and maintaining inter-vehicle safety distances.

Figure 7(A-c) shows successful obstacle navigation between buildings, with some UAVs utilizing pathways above and below obstacles, including the dynamic obstacle. Figure 7(A-d) demonstrates successful formation recovery after passing the buildings. Figure 7(B) presents the time history of distances between UAV1 and other UAVs, with vertical dashed lines corresponding to the scenarios shown in Figure 7(A). Results confirm consistent maintenance of safe distances exceeding 2 meters from UAV1.

The results from this section demonstrate our method's scalability with 3D maneuvers, successfully increasing the swarm size from 8 to 12 UAVs without additional complexity. This suggests that for larger spaces allowing larger barriers, our approach provides an effective solution for deploying any number of UAVs. (The complete simulation video is available in supplementary materials.)

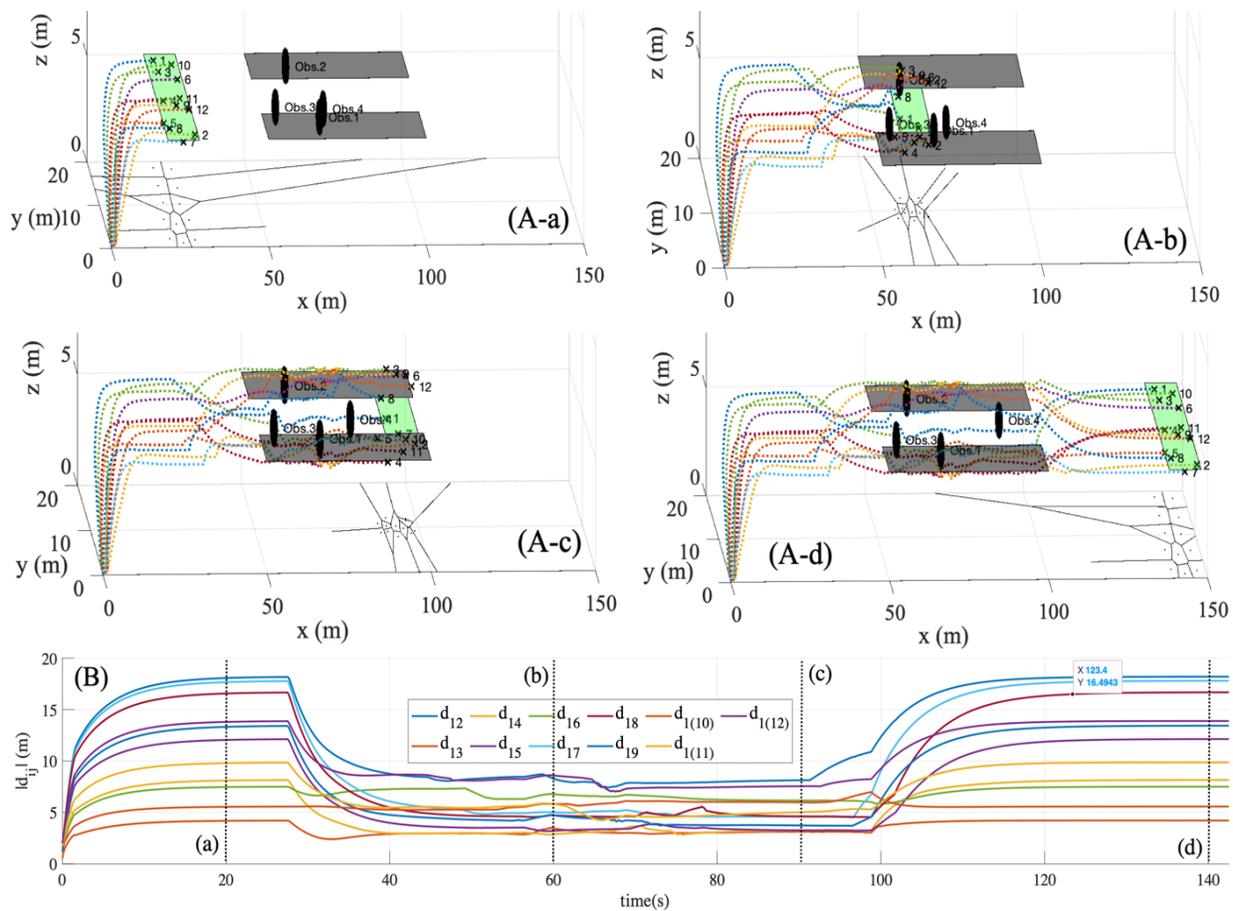

**Figure 7.** Simulation results for collision-free control scenario with nonplanar 3D obstacle avoidance maneuver

## CONCLUSION

This research addressed the challenges of implementing multi-UAV systems in urban environments containing both static and dynamic obstacles. Drawing inspiration from tilapia fish territorial behavior and pigeons' self-organized obstacle avoidance, combined with the probabilistic Lloyd's algorithm for CVT, we developed a nature-inspired collision-free formation control system with 3D obstacle avoidance capabilities for nonplanar multi-UAV missions. We validated our controller through three progressively complex scenarios: (1) Formation change between buildings without obstacles, (2) The same scenario with added static and dynamic obstacles, and (3) Large-scale implementation with 12 UAVs using full 3D maneuvers. Results demonstrated successful performance in obstacle avoidance, formation change, and formation recovery maneuvers. Future work could extend this framework to develop a learning-based neuromorphic digital-twin network for improved energy efficiency.

## REFERENCES

[1] J. Tang, G. Liu and Q. Pan , "A review on representative swarm intelligence algorithms for solving optimization problems: Applications and trends," *IEEE/CAA Journal of Automatica Sinica,* pp. 1627-1643, 2021.

[2] R. J. Amala Arokin Nathan , I. Kurmi and O. Bimber , "Drone swarm strategy for the detection and tracking of occluded targets in complex environments," *Communications Engineering,* vol. 2, no. 1, p. 55, 2023.

[3] J. Guo, J. Qi, M. Wang, C. Wu, Y. Ping, S. Li and J. Jin, "Distributed cooperative obstacle avoidance and formation reconfiguration for multiple quadrotors: Theory and experiment," *Aerospace Science and Technology,* vol. 136, p. 108218, 2023.

[4] S. Huang and R. S. H. Teo, "Computationally Efficient Visibility Graph-Based Generation Of 3D Shortest Collision-Free Path Among Polyhedral Obstacles For Unmanned Aerial Vehicles," in *2019 International Conference on Unmanned Aircraft Systems (ICUAS)*, 2019, (pp. 1218-1223).

[5] Y. I. Jenie, E. J. Kampen, C. C. de Visser, J. Ellerborek and J. M. Hoekstra, "Selective velocity obstacle method for deconflicting maneuvers applied to unmanned aerial vehicles," *Journal of Guidance, Control, and Dynamics,* vol. 38, no. 6, pp. 1140-1146, 2015.

[6] F. Bounini, D. Gingras, H. Pollart and D. Gruyer, "Modified artificial potential field method for online path planning applications," in *2017 IEEE Intelligent Vehicles Symposium (IV)*, (pp. 180-185), 2017.

[7] E. Soria, F. Schiano and D. Floreano, "Predictive control of aerial swarms in cluttered environments," *Nature Machine Intelligence,* vol. 3, no. 6, pp. 545-554, 2021.

[8] Y. Rasekhipour, A. Khajepour, S. K. Chen and B. Litkouhi, "A potential field-based model predictive path-planning controller for autonomous road vehicles," *IEEE Transactions on Intelligent Transportation Systems,* vol. 18, no. 5, pp. 1255-1267, 2016.

[9] Z. Yu, Z. Si, X. Li, D. Wang and H. Song, "A Novel Hybrid Particle Swarm Optimization Algorithm for Path Planning of UAVs," *IEEE Internet of Things Journal,* vol. 9, no. 22, pp. 22547-22558, 2021.

[10] H. Chu, J. Yi and F. Yang, "Chaos Particle Swarm Optimization Enhancement Algorithm for UAV Safe Path Planning," *Applied Sciences,* vol. 12, no. 18, p. 8977, 2022.

[11] Q. Yao, Z. Zheng, L. Qi, H. Yuan, X. Guo, M. Zhao, Z. Liu and T. Yang , "Path Planning Method With Improved Artificial Potential Field—A Reinforcement Learning Perspective," *IEEE Access,* vol. 8, pp. 135513-135523, 2020.

[12] Z. Pan, C. Zhang, Y. Xia, H. Xiong and X. Shao, "An Improved Artificial Potential Field Method for Path Planning and Formation Control of the Multi-UAV Systems," *IEEE Transactions on Circuits and Systems II: Express Briefs,* vol. 69, no. 3, pp. 1129-1133, 2022.

[13] G. Hao, Q. Lv, Z. Huang, H. Zhao and W. Chen, "Uav path planning based on improved artificial potential field method," *Aerospace,* vol. 10, no. 6, p. 562, 2023.

[14] F. Kong, Q. Wang, S. Gao and H. Yu, "B-APFDQN: A UAV path planning algorithm based on deep Q-network and artificial potential field," *IEEE Access,* vol. 11, pp. 44051-44064, 2023.

[15] H. T. Lin, I. G. Ros and A. A. Biewener, "Through the eyes of a bird: modelling visually guided obstacle flight," *Journal of the Royal Society Interface,* vol. 11, no. 96, p. 20140239, 2014.

[16] M. Huo, H. Duan, Q. Yang, D. Zhang and H. Qiu, "Live-fly experimentation for pigeon-inspired obstacle avoidance of quadrotor unmanned aerial vehicles," *Science China Information Sciences,* vol. 62, pp. 1-8, 2019.

[17] M. Zhang, K. Ma, L. Liu and W. Zhang, "Multiobjective Spherical Vector-Based Particle Swarm Optimisation for 3D UAV Path Planning," in *2024 6th International Conference on Data-driven Optimization of Complex Systems (DOCS)*, (pp. 427-432), 2024.

[18] M. Abdel-Basset, R. Mohamed, K. M. Sallam, I. M. Hezam, K. Munasinghe and A. Jamalipour, "A Multiobjective Optimization Algorithm for Safety and Optimality of 3-D Route Planning in UAV," *IEEE Transactions on Aerospace and Electronic Systems,* 2024.

[19] L. Ju, T. Ringler and M. Gunzburger, "Voronoi tessellations and their application to climate and global modeling," in *Numerical techniques for global atmospheric models*, Springer, 2011, pp. 313-342.

[20] T. Elmokadem and A. V. Savkin, "Computationally-efficient distributed algorithms of navigation of teams of autonomous UAVs for 3D coverage and flocking," *Drones,* vol. 5, no. 4, p. 124, 2021.

[21] J. Cortes, S. Martinez, T. Karatas and F. Bullo, "Coverage control for mobile sensing networks," *IEEE Transactions on robotics and Automation,* vol. 20, no. 2, pp. 243-255, 2004.

[22] R. Olafti-Saber, "Flocking for multi-agent dynamic systems: Algorithms and theory," *IEEE Transactions on automatic control,* vol. 51, no. 3, pp. 401-420, 2006.

[23] A. Cortes, S. Martinez and F. Bullo, "Spatially-distributed coverage optimization and control with limited-range interactions," *ESAIM: Control, Optimisation and Calculus of Variations,* vol. 11, no. 4, pp. 691-719, 2005.

[24] G. W. Barlow, "Hexagonal territories," *Animal Behavior,* vol. 22, pp. 876-IN1, 1974.

[25] B. Wie, "Attitude Dynamics and Control," in *Space Vehicle Dyamics and Control*, Amaerican Institute of Aeronautics and Astronautics, 1998, pp. 307-322.